\documentclass{article}

\usepackage{microtype}
\usepackage{graphicx}
\usepackage{booktabs} 

\usepackage{amssymb}
\usepackage{amsmath}
\usepackage{varwidth}
\usepackage{subfig}
\usepackage{multirow}

\usepackage{hyperref}

\usepackage[accepted]{icml2018workshop}

\icmltitlerunning{Airline Passenger Name Record Generation using Generative Adversarial Networks}

\begin{document}

\twocolumn[
\icmltitle{Airline Passenger Name Record Generation using Generative Adversarial Networks \\
           }



\icmlsetsymbol{t}{$\mp$}
\icmlsetsymbol{to}{$\dagger$}

\begin{icmlauthorlist}
\icmlauthor{Alejandro Mottini*}{t}
\icmlauthor{Alix Lh\'{e}ritier}{to}
\icmlauthor{Rodrigo Acuna-Agost}{to}
\end{icmlauthorlist}


\icmlcorrespondingauthor{Alejandro Mottini}{amottini@amazon.com}
\icmlcorrespondingauthor{Alix Lh\'{e}ritier}{alix.lheritier@amadeus.com}

\icmlkeywords{Synthetic Data Generation, Generative Adversarial Network, Passenger Name Records, Travel Industry}

\vskip 0.3in
]



\printAffiliationsAndNotice{*Work done at Amadeus SAS. $\mp$ Amazon, Alexa, Seattle, USA. $\dagger$ Amadeus SAS, Research, Innovation and Ventures Division, Sophia Antipolis, France}  

\begin{abstract}
Passenger Name Records (PNRs) are at the heart of the travel industry. Created 
when an itinerary is booked, they contain travel and passenger information. It 
is usual for airlines and other actors in the industry to inter-exchange and 
access each other's PNR, creating the challenge of using them without 
infringing data ownership laws. To address this difficulty, we propose a method 
to generate realistic 
synthetic PNRs using Generative Adversarial Networks (GANs). Unlike other GAN applications, PNRs consist of categorical and numerical 
features with missing/NaN values, which makes the use of GANs challenging. We propose a solution based on Cram\'{e}r GANs, categorical 
feature embedding and a Cross-Net architecture. 
The method was tested on a real PNR dataset, and evaluated in terms of distribution matching, memorization, 
and performance of predictive models for two real business problems: client segmentation and passenger nationality prediction. 
Results  show that the generated data matches well with the real PNRs without 
memorizing them, and that it can be  used to train models for real business 
applications.
\end{abstract}

\section{Introduction}

Passenger Name Records (PNRs) store the travel information of an individual or group of passengers travelling together. 
They are generated when a travel reservation is made, and stored by airlines, travel agencies and/or Global Distribution Systems (GDS).




PNRs are important to all travel providers, and are routinely used for business applications such as client segmentation and adaptive product pricing \cite{usos_airlines}. However, access to PNRs is severely limited.  Within airlines, passenger privacy concerns hinders them from developing data-driven commercial applications. In addition, although GDS store in their servers PNRs for hundreds of airlines and travel providers, this data does not belong to them and their use is limited due to ownership issues. Moreover, similarly to what happens in the medical and financial fields, the inability to share data makes the collaboration with external partners (academia, start-ups, etc) difficult. This is particularly true in the European market, where the new EU General Data Protection Regulation (GDPR) \cite{gdpr} introduced strict data privacy regulations. Under this new data regulation, data must be deleted after its original purpose passed.



One way of addressing these difficulties is by using synthetic data, which 
should have the original data structure and follow its distribution 
sufficiently well to meet the goals of the application at hand, without 
memorizing the original data.
Our goal is to generate realistic data 
that can be used for different data-driven 
business applications (testing of production pipelines, internal training, 
etc)  without data ownership concerns.

Generative adversarial networks (GANs) \cite{gansIan2014} is a framework for learning deep generative models that has been gaining popularity. It has been successfully applied to many fields including images \cite{gan_img2}, natural language \cite{gan_language} and anomaly detection \cite{gan_medic}. 

In contrast to most  previous GAN applications, PNR data consists of categorical and numerical 
features with missing values, which makes the use of GANs challenging. In this paper we propose to use Cram\'{e}r GANs \cite{cramer_Gan} with a generator/critic architecture that combines feed-forward layers with the Cross-Net architecture \cite{crossNet} and  uses an input embedding layer for the categorical features.

The method was tested on a real PNR dataset comprising international trips in different  airlines and regions of the world. 
A special emphasis is put on the evaluation of the generated data, an important 
problem in the GAN literature \cite{gan_mmdDiv,lopez2016revisiting}.

Most validation methods are designed for the image domain, and either rely on the direct evaluation of the generated samples or on some perceptual metric such as the Fr\'{e}chet Distance \cite{gan_fid_dist} or the Inception score \cite{gan_img2}. Therefore, these methods are not directly applicable to our data. In addition, manual evaluation of the generated PNRs would be difficult and cumbersome. Instead, we propose several methods better adapted to our application. Most importantly, the synthetic data is used to train classification models 
associated to two important data-driven business applications in the travel 
industry: client segmentation and nationality prediction. Classification models 
are trained on real or synthetic data, and evaluated on a test set of real samples. The difference in performance between a model trained on real data and another trained on synthetic data is used to assess the quality of the generated data.




Finally, we compare the proposed approach with other state of the art methods, including Wasserstein GANs \cite{wgan_imprv} with fully connected generator/critic architectures.



\section{Related Work}
\label{sec:rwork}

There is a growing interest in the generation of realistic synthetic data. The challenge consists in generating data that reproduces both the structural and statistical properties of the original data sufficiently well for the desired application, but whose values are not obtained by direct observation of the real generative process. 

To produce synthetic data, generative models are trained on real samples drawn from a given distribution, and learn to approximate it. The resulting distribution can be learnt explicitly, or be implicitly encoded in the generator \cite{ian_tutorial}.  Many generative models exist in the literature, two of the most popular being Variational auto-encoders (VAE)  \cite{vae} and Generative adversarial networks (GANs)  \cite{gansIan2014}. The latter has gained a considerable amount of interest recently, producing state-of-the-art results in terms of quality of the generated samples \cite{gan_img2}


There are many GAN  models designed for specific applications in the literature, but none that target data that combines numerical and categorical features that contains missing/NaN values. In \cite{medical_record}, the authors propose a model that combines  auto-encoders and GANs to generate Electronic Health Records. The method can handle both binary and count variables. The authors use the original GAN formulation \cite{gansIan2014} that is particularly hard to train and susceptible to the ``mode collapse" problem \cite{gan_img2}. In addition, \cite{gan_senseGen} uses GANs to generate synthetic sensor data comprised of  accelerometer traces collected using smart-phones.  The proposed architecture uses a multilayer Long-Short-Term-Memory (LSTM) networks and a Mixture Density Network as generator, and a second LSTM network as discriminator. The model is able to produce realistic sequences, but is designed to work on sequential 1D datasets and is thus not appropriate for our  dataset.

\section{PNR Data}
\label{sec:pnr}

PNRs are created at reservation time  by airlines and/or travel providers, and are then stored in the airline's  or GDS data centers. The records must be deleted three months after the trip has taken place.

Raw PNRs are created in the EDIFACT format, a semi structured messaging format
used throughout the airline industry.  

The main fields present in the records are the ones describing the trip. These are present in all PNRs and contain information such as origin/destination airports and the booking agency's code (referred to as Office Id). Additional elements such as personal
information of the passengers (e.g., nationality), payment information (e.g., price), and other details can be included. In total, a PNR can contain up to 60 fields, although typically, fewer are used in practice \cite{iata}.  These fields can contain numerical, categorical and date data, as well as missing or NaN values.

Given that most data-driven applications build on top of PNRs only need certain data fields, we extract them from the original EDIFACT files and transform the PNRs into a structured dataset. In particular, we are interested in the fields commonly used for the two business applications considered in the evaluation: business/leisure segmentation and passenger nationality prediction (see Section \ref{sec:validation}). It should be noted that these features do not contain any information that can identify a particular passenger. This is why our main concern is not to generate synthetic data that protects the privacy of individual passengers, but rather, to generate realistic data that we can use for different business applications (passenger segmentation, testing of production pipelines, internal training, etc) and shared with external partners without legal constraints. A complete list of the selected features is presented in Table \ref{tbl:features}.

\begin{table}[t]
  \caption{Type (categorical, binary or numerical) and range/cardinality of the features used to represent a PNR.}
  \label{tbl:features}
  \vskip 0.15in
  \begin{center}
  \begin{small}
\begin{sc}
  \begin{tabular}{lcr}
    \toprule
    Feature & Type & Range/Card. \\
    \midrule
Country Origin  & Cat. & 81\\ 
Country Destination & Cat. & 95\\ 
Country Office Id. & Cat. & 65\\ 
Stay Saturday & Binary & $\{$0,1$\}$\\  
Purchase Anticipation & Num. & [0,364] \\  
Number Passengers & Num. & [1,9] \\  
Stay duration days& Num. & [0,90] \\  
Gender & Binary & $\{$0,1$\}$\\  
PNR With Children & Binary & $\{$0,1$\}$\\  
Age (years)& Num. & [0,99] \\  
Nationality & Cat. &  76\\ 
Business/Leisure & Binary & $\{$0,1$\}$\\  
  \bottomrule
\end{tabular}
\end{sc}
\end{small}
\end{center}
\vskip -0.1in
\end{table}

\section{Method}
\label{sec:cmodel}

In this section we briefly present the original GAN framework and some recent improvements proposed in the literature. Finally, we explain in detail our generation architecture.

\subsection{Generative Adversarial Network}
\label{ssec:gan}

In the original GAN formulation \cite{gansIan2014}, a generative model (G) and 
a discriminative model (D) are trained simultaneously with conflicting 
objectives.  Given a training set of real samples, G is trained to approximate 
the real data distribution, while D is trained to discriminate between real and 
synthetic samples.

Formally, GAN solves the following min-max game:
\begin{equation}
\min_G \max_D \mathcal{L}(G,D)
\label{eq:gan1}
\end{equation}
where:
\begin{equation}
\begin{split}
\mathcal{L}(G,D) =  \mathbb{E}_{X \sim P_{data}} [\log (D(X))]\ + \\ \mathbb{E}_{Z \sim P_{Z}} [log(1-D(G(Z))]
\end{split}
\label{eq:gan2}
\end{equation}

In the previous equation, $P_{data}$ is the real training data distribution from which samples $x$ are drawn, and $P_{Z}$ is a distribution from which noise input vectors $z$ are drawn. G is a mapping from $z$ to $x$ space, while D maps an input $x$ to a scalar value that represents the probability of $x$ being a real sample. In practice, both G and D are neural networks whose architecture depends on the application and are trained alternately using a gradient descent based method. At the end of the optimization process, if an equilibrium is reached, G learns a distribution $P_{model}$ that closely resembles  the real data distribution $P_{data}$, and D is unable to distinguish the real from the generated samples (i.e., output  probability $\frac{1}{2}$).

Although highly successfully, GANs are known to be difficult to train, due in part to the choice of the divergence measure used for the comparison of the distributions. This choice has been the subject of recent work, where a number of divergences have been proposed, including Maximum Mean Discrepancy (MMD) \cite{gan_mmdDiv}, the Wasserstein metric \cite{wgan,wgan_imprv}  and  the Cram\'{e}r  distance \cite{cramer_Gan}.


More precisely, \cite{wgan} proposes to use the Wasserstein distance instead of the Jensen-Shannon divergence (JSD).  This distance has better theoretical properties than JSD, particularly when working with distributions with non-overlapping support.


Wasserstein GANs (WGANs) are considered to be easier to train that the original GANs, produce better sample quality and have already been successfully applied in the literature \cite{gan_language, gan_language2}. Nevertheless, in has been shown \cite{cramer_Gan} that they can present problems when optimized with Stochastic gradient descent (SGD) based methods. The problem arises when estimating the Wasserstein metric from samples, which might yield biased gradients  and converge to a wrong minimum. This problem is referred to as biased gradients.

To overcome this difficulty, \cite{cramer_Gan} proposes to use the Cram\'{e}r Distance instead, which has the same good properties of Wasserstein, while providing unbiased sample gradients. More precisely, the authors use the multivariate extension of the Cram\'{e}r distance, the energy distance  \cite{cremer_dist}:
\begin{equation}
\begin{split}
\mathcal{E}(P,Q) := \mathcal{E}(X,Y) := 2 \mathbb{E} ||X-Y||_2 + \\
- \mathbb{E}||X-X^{'}||_2 -    \mathbb{E}||Y-Y^{'}||_2 
\end{split}
\label{eq:cramer}
\end{equation}
where $X, X^{'}, Y, Y^{'}$ are independent random variables distributed according to $P,Q$, respectively. The energy
distance can also be written in terms of a difference of expectations:
\begin{equation}
\begin{split}
\mathcal{E}(X,Y) := \mathbb{E} f^*(X) - \mathbb{E} f^*(Y)   \\
f^*(x) := \mathbb{E} ||x-Y^{'}||_2  -   \mathbb{E} ||x-X^{'}||_2
\end{split}
\label{eq:cramer2}
\end{equation}

In practice, the authors combine the energy distance with a transformation function $h: R^d \rightarrow R^k$, where $d$ is the dimensionality of the input samples and $k$ is a hyper-parameter. This transformation, implemented with a neural network,  is used to define D (referred to as the critic in the WGAN and Cram\'{e}r GAN literature) and is meant to map the input into a space where the discrimination between the distributions is easier. Thus, $f^*(x)$ is replaced with:
\begin{equation}
\begin{split}
f(x) := \mathbb{E}_{Y^{'} \sim Q} ||h(x) - h(Y^{'})||_2 +    \\
 - \mathbb{E}_{X^{'} \sim P} ||h(x) - h(X^{'})||_2
\end{split}
\label{eq:cramer3}
\end{equation}

Therefore, D only has trainable parameters inside the transformation network $h$. 


Finally, given $X,Y$ real and generated samples drawn from $P_{data}$, $P_{model}$ respectively, the losses used by the Cram\'{e}r-GAN can be defined as:
\begin{equation}
\begin{split}
\mathcal{L}_{G} (X,Y) = \mathbb{E}_{X \sim P_{data}} [f(X)] - \mathbb{E}_{Y \sim P_{model}}[f(Y)]   \\
\mathcal{L}_{D} (X,Y)= -\mathcal{L}_{G} (X,Y) + \lambda GP 
\end{split}
\label{eq:cramer4}
\end{equation}
%


where   $\lambda GP$ is referred to as the gradient penalty term, and its purpose is to penalize functions that have a high gradient. For a detailed description of the method please refer to  \cite{cramer_Gan}. The training procedure is otherwise  similar to the one described in \cite{wgan_imprv}, which consists of  performing $n_{critic}$ critic updates per generator update.







\subsection{PNR Generation with Cram\'{e}r GANs}
\label{ssec:pnrgan}

The first step of the generation process is data pre-processing.  PNRs have both numerical and categorical features. In the case of the numerical features, the pre-processing is straight-forward (except for the treatment of missing/NaN values, which is described later in this section) and only involves scaling them to the $[0,1]$ interval. On the other hand, categorical features require a special treatment. 

\begin{figure}[ht]
\vskip 0.2in
\begin{center}
\centerline{\includegraphics[angle=90, scale=0.37]{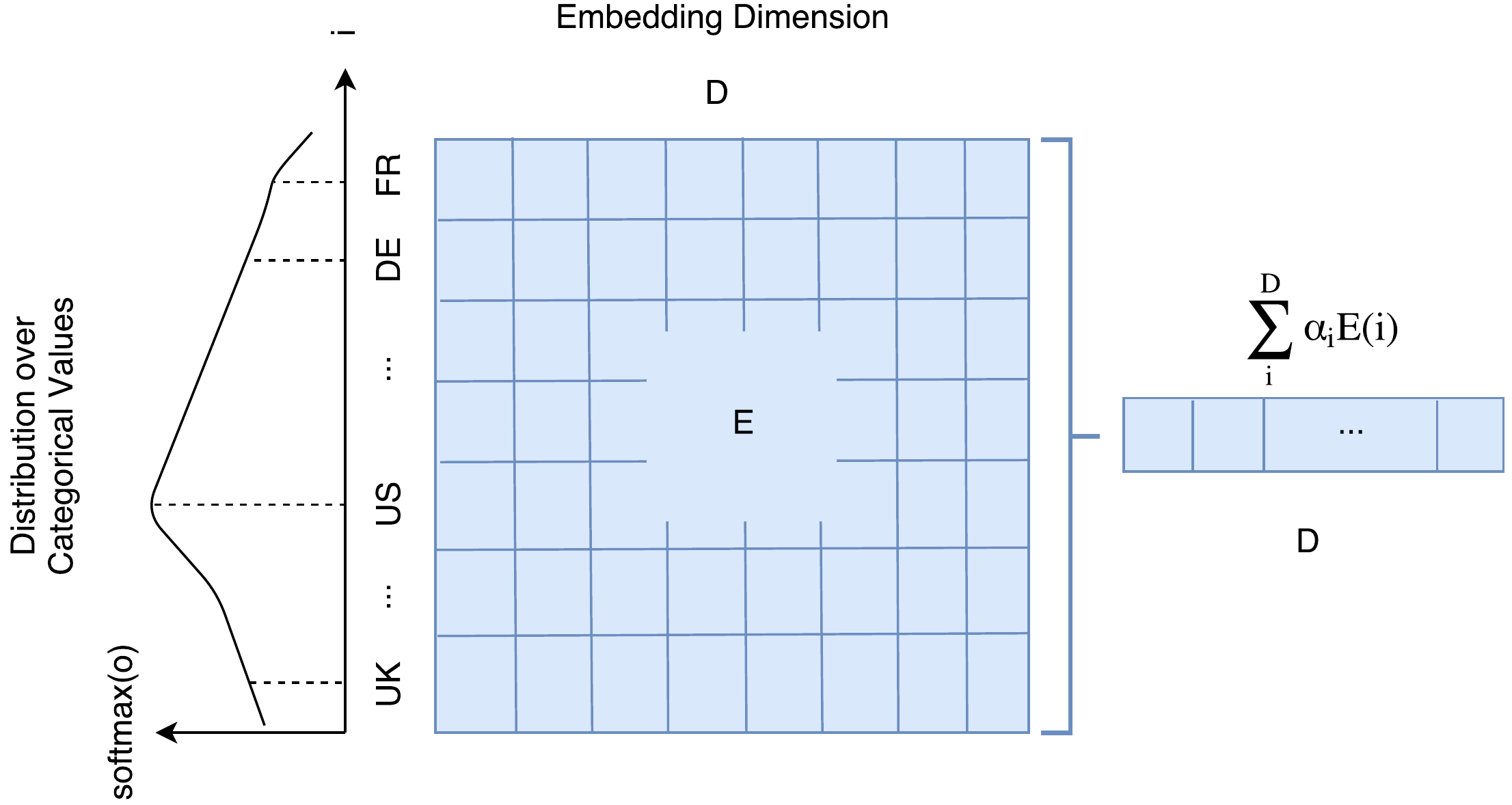}}
\caption{Weighted average of the embedded representation of each discrete token of the Country Origin categorical feature.}
\label{fig:embeding_promedioMatrix}
\end{center}
\vskip -0.2in
\end{figure}

The use of discrete data  with GANs is not trivial, and has been the subject of much study recently.  Since the generator must be differentiable, it cannot generate discrete data (such as one-hot encoded values or character representations) \cite{gansIan2014}. One way to tackle this problem is to encode discrete columns into numerical columns \cite{gansIan2014} . More recently, authors have used the Gumbel-Softmax \cite{gan_gumbel} to approximate the gradients  from discrete samples. Finally,  within the framework of language generation, \cite{gan_language} propose a GAN architecture that uses RNNs for both the generator and the discriminator, and to use the weighted average of the embedded representation of each discrete token instead of simply the embedding of the most probable token. The embedding layer is shared between G and D. This continuous relaxation enforces a fully-differentiable process (see Figure \ref{fig:embeding_promedioMatrix}).

Through experimentation we have determined that this last method produces the best results for our problem and data. We believe this is in part due to the increase representational power provided by the embedding layers. This is particularly  true when comparing this approach with the simple encoding of the categorical values into a numerical column, where all the categorical information has to be condensed into a single numerical value. 

\begin{figure}[ht]
\vskip 0.1in
\begin{center}
\centerline{\includegraphics[width=\columnwidth]{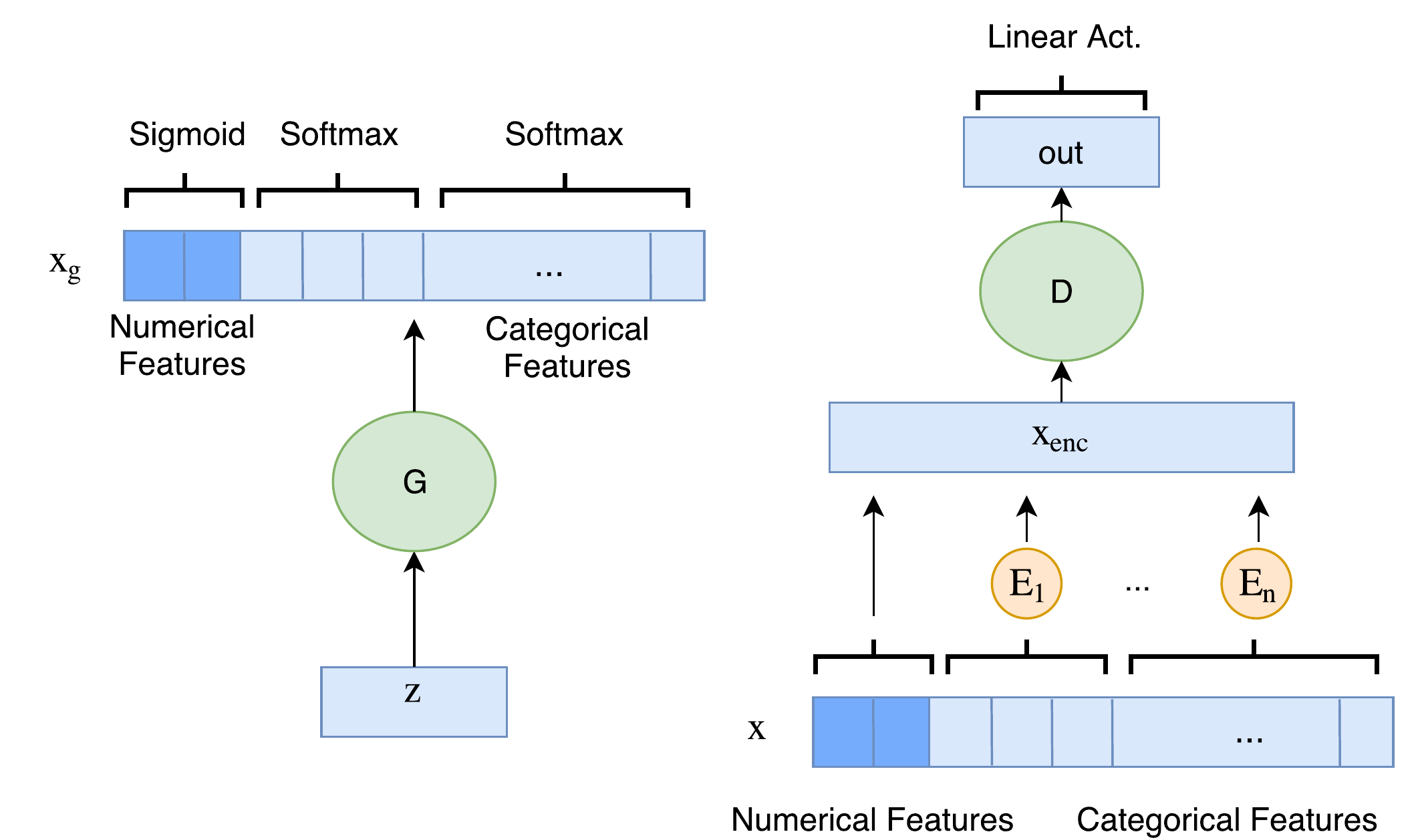}}
\caption{G and D inputs and outputs.}
\label{fig:embeding_promedio}
\end{center}
\vskip -0.1in
\end{figure}

Therefore, categorical features are one-hot-encoded at this stage. It should be noted that unlike \cite{gan_language}, we use a different embedding layer per categorical feature. Moreover, in our case, the embeddings are only needed in D. Our generator produces soft one hot encoded values (see Figure \ref{fig:embeding_promedio}). 




The final pre-processing step consists of dealing with missing or NaN values. For the categorical features, these cases are simply replaced with a new level "UNK". For numerical features, a two step process is performed. First, the missing values (or NaN) are filled in by replacing them with a  random value taken from the same column. Then, a new binary column is added, whose values are 1 for all the filled-in rows, and 0 otherwise. One such column is added per numerical column with missing values. These auxiliary binary columns are treated as categorical columns, and encoded using the same processes detailed before.





Most current works on GANs use either image or sequence data. Therefore, the generator's and discriminator's architecture usually consists of  layers of Convolutional (CNN) \cite{gan_img2} or Recurrent Neural networks  (RNN) \cite{gan_language2}. However, our data presents a structure that is more compatible with feed-forward neural networks (FNN). 

\begin{figure}[ht]
\vskip 0.1in
\begin{center}
\centerline{\includegraphics[width=\columnwidth]{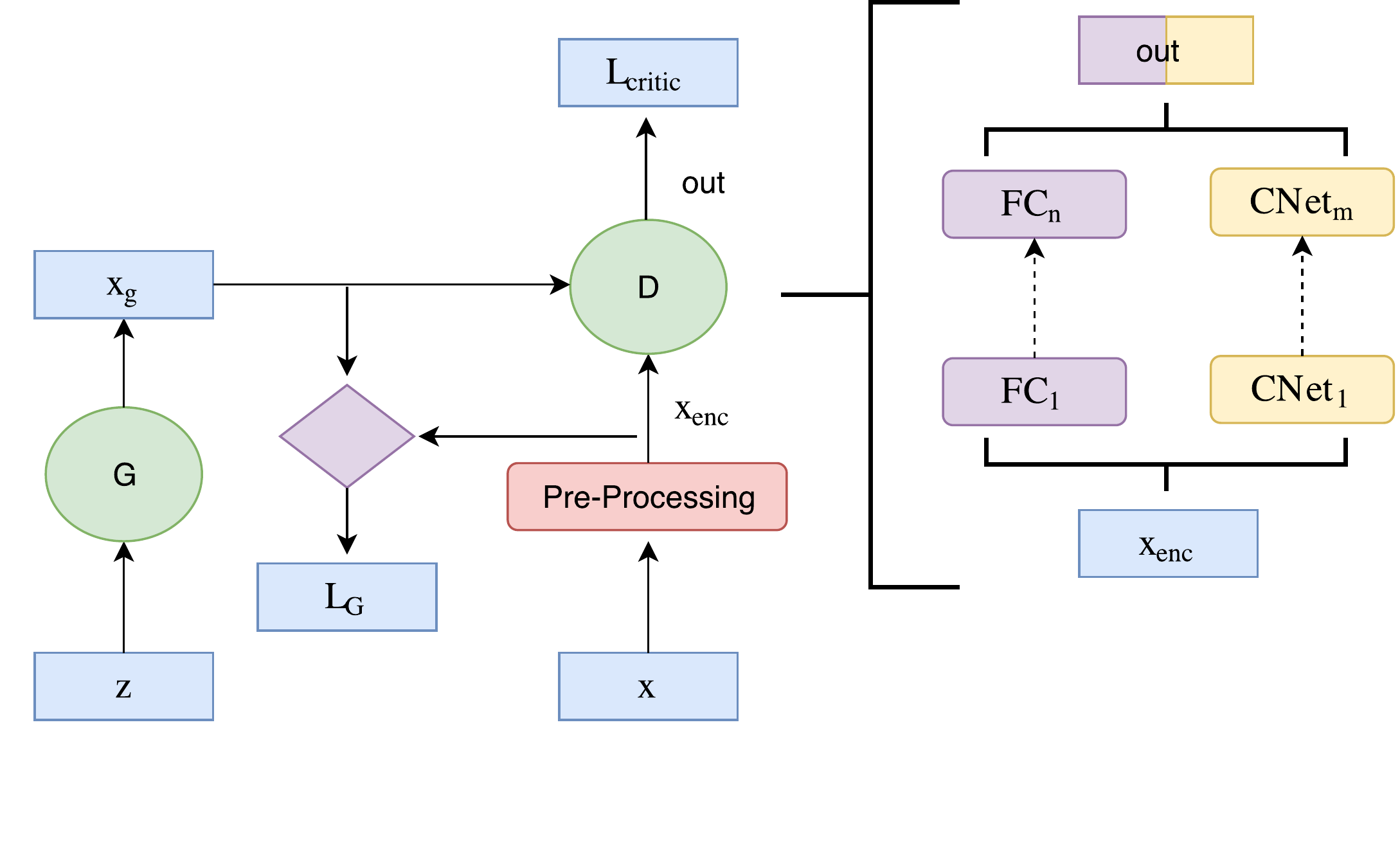}}
\caption{Network architecture.}
\label{fig:gan_schema}
\end{center}
\vskip -0.1in
\end{figure}

Multilayer FNNs are able to learn complex feature interactions. Nevertheless, they might fail to efficiently learn cross feature interactions, which have been shown to improve models' performance in similar problems \cite{crossNet}. To address this shortcoming, we propose an architecture for G and D as described in Figure \ref{fig:gan_schema}. Both G and the transformation $h$ (used to define D) are composed of fully connected layers and cross-layers \cite{crossNet}, a new type of architecture that explicitly produces cross feature interactions. By stacking $N$ cross layers, we are able to automatically compute up to $N$-degree feature interactions in an efficient manner.




Leaky ReLU activations are used for all but the last layer of G, which uses a Sigmoid activation on the numerical features  and Softmax activations for the categorical ones (one per feature). Thus, G generates distributions over the categorical values of each discrete feature.

D receives real and generated samples composed of both  $[0,1]$ numerical features and categorical values distributions for the discrete features (either one-hot vectors for real samples or Softmax distributions for synthetic ones).  These categorical values distributions are projected into lower dimensional dense representations using embeddings. In the case of the one-hot encoded samples, the embeddings act as simple lookup tables. On the other hand, the soft distributions over discrete values produce a weighted average of the  embedding matrix rows, where each row represents the embedding of each categorical value in the new lower dimensional space.  Each discrete feature has a separate embedding matrix, which is initialized randomly and learned jointly with all other model parameters through back-propagation. The process is illustrated in Figure \ref{fig:embeding_promedio}. D also uses Leaky ReLU activations in all but the last layer, which consists of a dense layer with linear activation. As it is usual with Wasserstein and Cram\'{e}r GANs, no batch normalization nor dropout are used. 

\begin{table}[t]
 \caption{Hyper-parameters used for the PNR generation.}
  \label{tbl:params}
  \vskip 0.15in
  \begin{center}
  \begin{small}
\begin{sc}
  \begin{tabular}{lr}
     \toprule
    Name & Value\\
    \midrule
Opt. algorithm & Adam \\  
Learning rate  &  0.0001 \\  
Batch size & 128 \\
z & $U[0,1]^{12}$ \\  
G. & (64,128) \\  
h & (64,128,128)  \\  
$\lambda$ & 10.0  \\  
$n_{critic}$ & 5 \\
Cross layers & 2 \\
  \bottomrule
\end{tabular}
\end{sc}
\end{small}
\end{center}
\vskip -0.1in
\end{table}

At the end of the process, the generated data is post-processed. Numerical features are returned to their original range, while categorical columns are transformed back to the discrete domain by assigning the value with the highest probability of the softmax output. Finally, the auxiliary binary columns associated to missing values in the numerical columns are used to replace the indicated values in those columns with NaN. These auxiliary binary columns are then removed.

\section{Validation}
\label{sec:validation}

The model was validated on a real dataset that contains PNRs sampled  from international trips in several airlines and regions of the world. 
The raw PNRs were parsed and transformed from the EDIFACT format into a structured dataset using a Spark
cluster. The resulting dataset\footnote{Data and code available for download here: \url{https://www.dropbox.com/sh/11c5mo0f7g42wgs/AACm7xaRdyZdzMo2NgCPIrPja?dl=0}\label{footnote}} consists of approximately one hundred thousand records contains both numerical and categorical features (see Table \ref{tbl:features}). The dataset is separated into a training and a test set. The generated data is compared with the real test set for all the different evaluations. 

The proposed model was implemented using Tensorflow and Scikit-learn and is available for download\textsuperscript{\ref{footnote}}. 




To determine the quality of the generated data, several evaluations are proposed, which are detailed in the following sections.





\subsection{Distribution matching}

In order to assess the quality of the results, the simplest method is to  
compare the univariate empirical distribution of each variable between real 
and synthetic data. 

To get a multivariate measure of how much
two distributions $P_X$ and $P_Y$ differ, we use the Jensen-Shannon divergence 
(JSD)
\begin{equation}
\text{JS}(P_X||P_Y) \equiv 
\frac{1}{2}\left(D_\text{KL}(P_X||\bar{P}) + 
D_\text{KL}(P_Y||\bar{P})\right)
\end{equation}
where $\bar{P}\equiv\frac{1}{2}(P_X+P_Y)$ and $D_\text{KL}$ denotes the 
Kullback-Leibler divergence (KLD).
A practical interest of this divergence is that, unlike the 
KLD, it is symmetric and its value is bounded between 0 and 1.   

To go further in this analysis and understand in more detail how the distributions differ, we use the divergence decomposition proposed in 
\cite{cazals2015beyond}. 
The mixture $\bar{P}$ implicitly defines a supervised learning problem with 
features $Z\sim\bar{P}$ and label $L\sim\text{Bernoulli}(1/2)$ indicating from 
which original distribution (i.e. $P_X$ or $P_Y$ ) an instance of $Z$ is 
obtained.
Then, the \emph{local  
discrepancy}, whose expected value over $Z$ correspond to the JSD, is defined as
\begin{equation}
\delta(z)  \equiv D_\text{KL}(\Pr(L=\cdot|z)||1/2).
\end{equation}
This decomposition is first  estimated in the input space, and then 
the  points  are projected in 2D using Multidimensional Scaling (MDS) 
to visualize them. 
The estimator uses $k$-nearest neighbors and requires data lying in an 
Euclidean space and, therefore, a 
transformation is needed for categorical variables. 
For this, we use the critic's $h$ function to embed all the data points into an 
Euclidean space.

Finally, it is crucial to assess the distribution matching in the  
original feature space. As proposed in \cite{lopez2016revisiting}, we 
train a classifier to discriminate between real and generated samples (e.g., labeling the real 
samples as 0 and the synthetic ones as 1). Under the null hypothesis of equal distributions, the expected accuracy is 1/2, 
a higher accuracy supporting the alternative hypothesis. Therefore, we can use 
the accuracy yielded by a classifier that properly handles categorical and numerical features (e.g., 
Random Forests)  as a measure of discrepancy 
in the original feature space. 

\subsection{Memorization}

To determine if the generative model is learning the original data distribution and not simply memorizing and reproducing the 
training data, we calculate the Euclidean  distances between each 
generated point and its nearest neighbour in the training and test sets. We then 
compare the distributions of distances using the Kolmogorov-Smirnov (KS) 
two-sample test to determine if they differ. We also consider the one-sided 
Bayesian Wilcoxon signed-rank test \cite{baysianW} to determine if the 
generated  points are closer to the training set than the test set.
Since our data points consist of both numerical and categorical 
features, we use, as before, the output of the critic to represent the samples 
in a numerical space. 




\subsection{Validation with Predictive models}
Importantly, we are  interested in determining if our 
approach is able to generate realistic PNR data that could be used in real 
data-driven business applications without compromising data ownership laws. In particular, we considered two applications: business/leisure segmentation \cite{clinet_segmentation} and nationality prediction \cite{pnr}.

In the first case, the objective is to determine to which segment a passenger 
belongs to. In the airline industry, there are two classical segments: business 
and leisure. Business passengers are product oriented without much consideration for the 
price. Leisure passengers are price sensitive customers.

On the other hand, knowing in advance the passenger's nationality is of particular 
importance to the airports and airlines, and is routinely used in different 
applications such as adaptive product pricing \cite{usos_airlines}. This 
attribute is usually predicted using machine learning methods or estimated at 
airports using surveys.

Both passengers segment and nationality are present in our training PNR dataset, but are not commonly present in most PNRs. 


We evaluate the quality of the generated PNRs by 
comparing the performance of classifiers trained on real vs synthetic data, 
and evaluated on a test set of real samples. We argue that if the model's 
performance does not degrade significantly when doing this cross evaluation, 
the generated data is good enough to be used for these (and potentially other) 
applications.

\subsection{Comparison with Alternative Methods}
\label{ssec:alternativeMethodss}

We have compared our method (referred to as CrGAN-Cnet in the Result section) against alternative approaches. First, we consider the same CrGAN-Cnet model but without using Cross-Networks in the generator/critic (generator/critic consist of only fully connected layers).  We refer to this method as CrGAN-FC. We also test our approach against a Wasserstein GAN using the same embedding approach  described in  Section \ref{ssec:pnrgan} and a generator/critic architecture of fully connected layers (referred to as WGAN-FC). 

In addition, as an alternative to using embeddings for the categorical features, discrete columns are encoded into numerical $[0,1]$ columns  using the following method. Discrete values are first sorted in descending order based on their proportion in the dataset. Then, the $[0,1]$ interval is split into sections $[a_c,b_c]$ based on the proportion of each category $c$.  To convert a discrete value to a numerical one, we replace it with a value sampled from a Gaussian distribution centered  at the midpoint of $[a_c,b_c]$ and with $\sigma=\frac{b-a}{6}$ as proposed in \cite{mit_genenration}. Both Cram\'{e}r  and Wasserstein GANs are tested with this approach (CrGAN-Num and WGAN-Num respectively). It should be noted that the generator's final activation function is changed to sigmoid in order to force the  generator to produce data that is always in the  $[0,1]$ range.

Due to computational constraints, we have not performed an extensive hyper parameter tuning of the methods. The parameters were kept constant for all models during the evaluation (see Table \ref{tbl:params}).


It should be noted that not all the evaluation methods proposed in Section \ref{sec:validation} are applicable to these alternative approaches, and are therefore excluded during the comparison of the different generative models.

\subsection{Results}
\label{ssec:results}

\begin{figure}[ht]
\vskip 0.1in
\centering
\subfloat[Business passengers]{\includegraphics[scale=0.5]{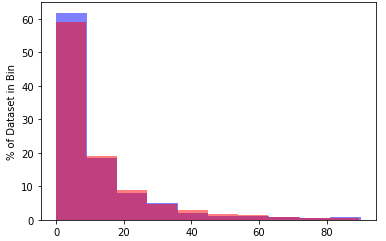}}
\subfloat[Leisure passengers]{\includegraphics[scale=0.5]{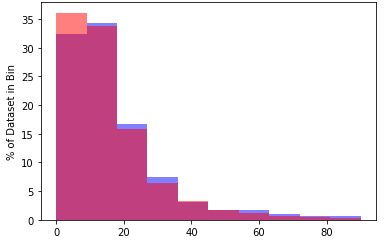} }
\caption{Univariate distributions evaluation (CrGAN-Cnet). Passenger's stay duration histogram comparison (real dataset in red, generated dataset in blue, superposition in pink) for business (a) and leisure (b) passengers.}
\label{fig:stay_hist}
\vskip 0.1in
\end{figure}

We start by evaluating our approach (CrGAN-Cnet) in depth using all the validation methods described before. First, we compare the univariate distributions between the real and simulated data.  Figure \ref{fig:stay_hist} and Table \ref{tbl:age_description} present results for the passenger's stay duration and age features. One can clearly see that the model produces data where the individual features follow the real distributions.


\begin{table}
  \caption{Univariate distributions evaluation. Passenger's age description of the real and generated datasets (CrGAN-Cnet).}
  \label{tbl:age_description}
   \begin{center}
  \begin{small}
\begin{sc}
  \begin{tabular}{lcr}
    \toprule
     & Real & Generated\\
    \midrule
Mean & 46.9 & 43.9\\  
Std  &  13.7 & 14.2\\  
Min & 0.01 & 3.3\\
25\% & 36.9 & 34.2\\  
50\% & 46.2 & 42.8\\  
75\% & 56.2  & 52.3\\  
Max & 99.0  & 96.7\\  
  \bottomrule
\end{tabular}
\end{sc}
\end{small}
\end{center}
\vskip -0.1in
\end{table}

\begin{figure}[ht]
\vskip 0.1in
\centering
\subfloat[JSD=0.38]{\includegraphics[scale=0.27]{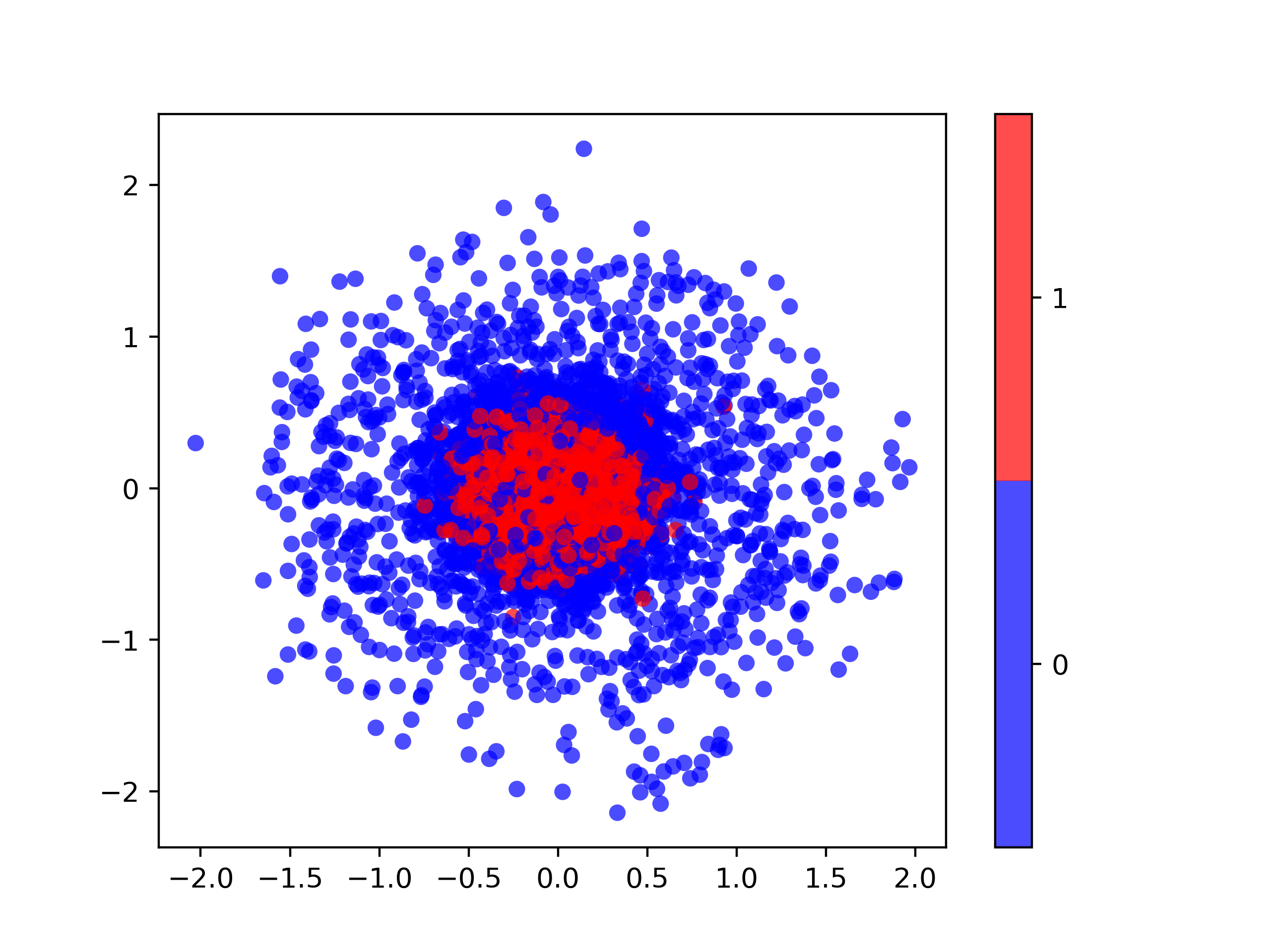}
	\includegraphics[scale=0.27]{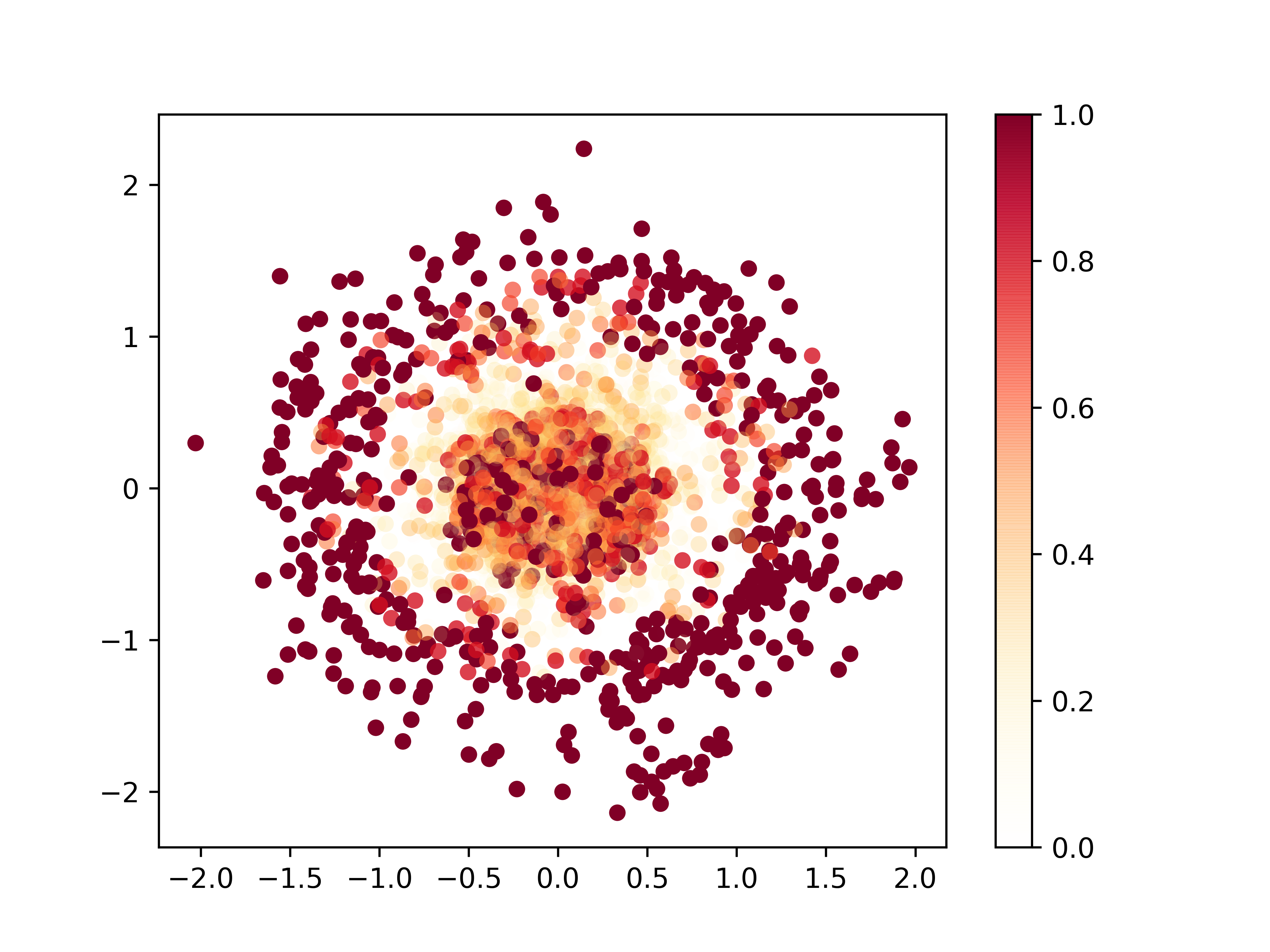}
\label{fig:huevos1}}
\\
\subfloat[JSD=0.21]{\includegraphics[scale=0.27]{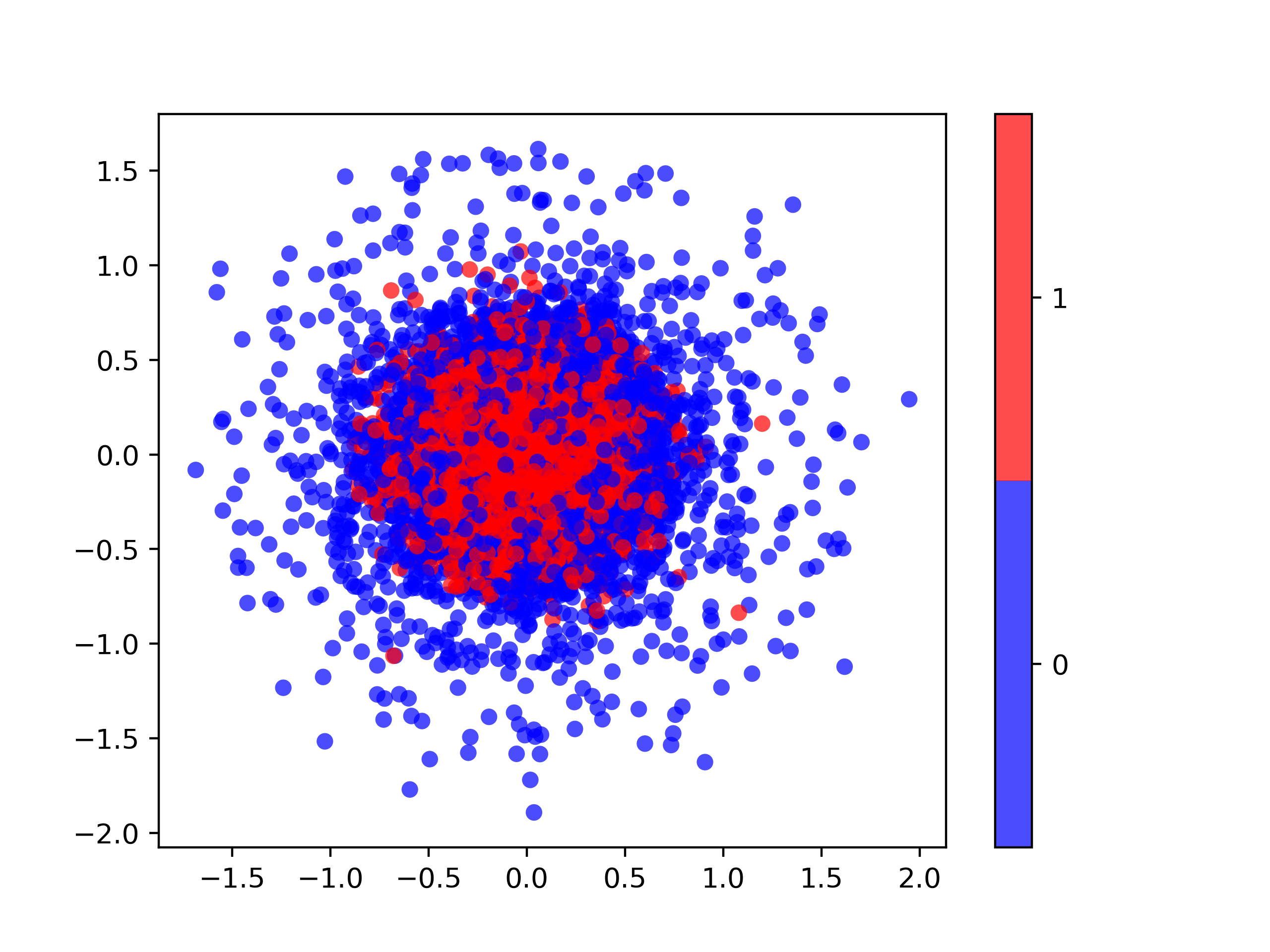}
	\includegraphics[scale=0.27]{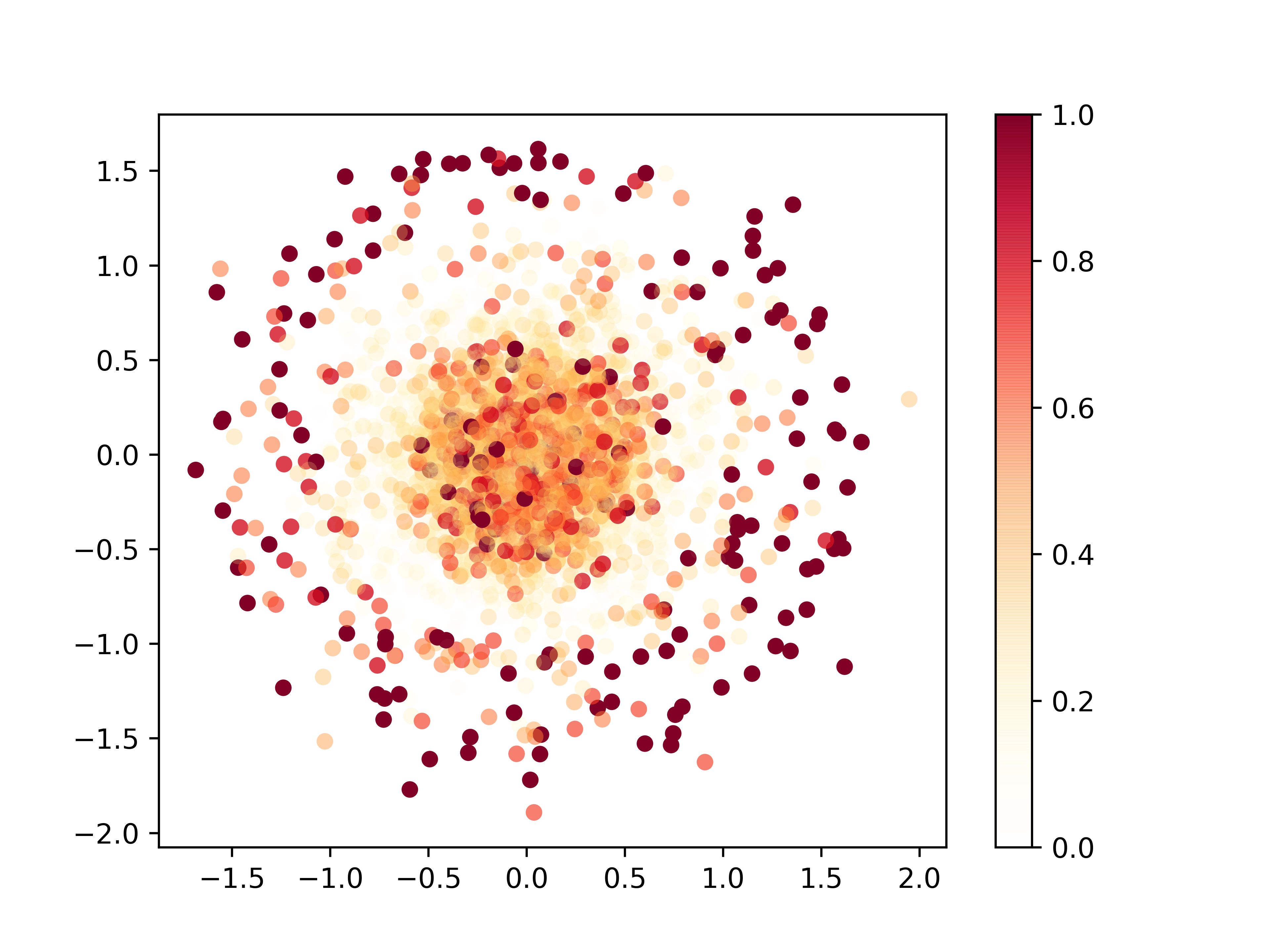}
\label{fig:huevos2}}
\\
\subfloat[JSD=0.08]{\includegraphics[scale=0.27]{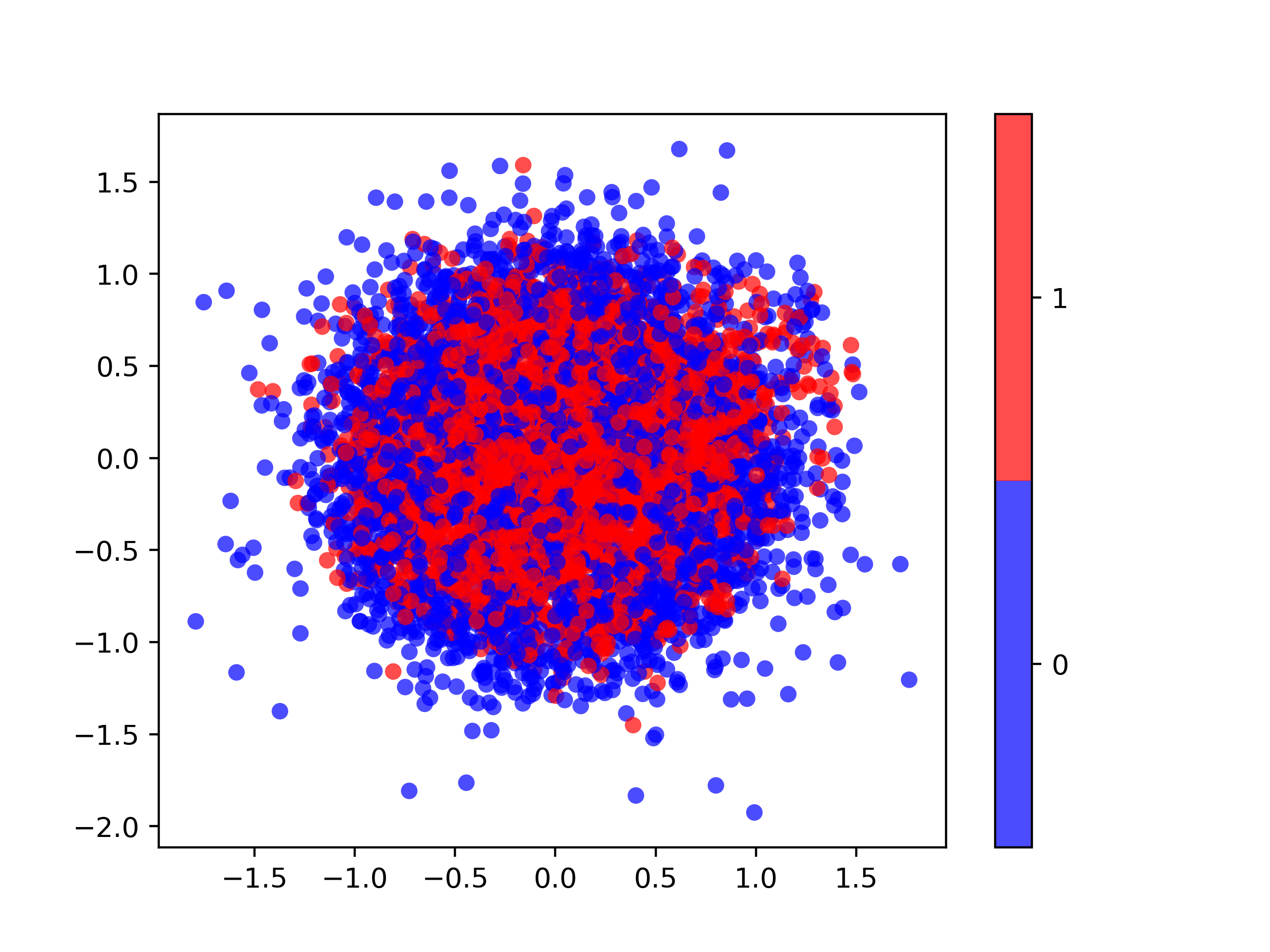}
	\includegraphics[scale=0.27]{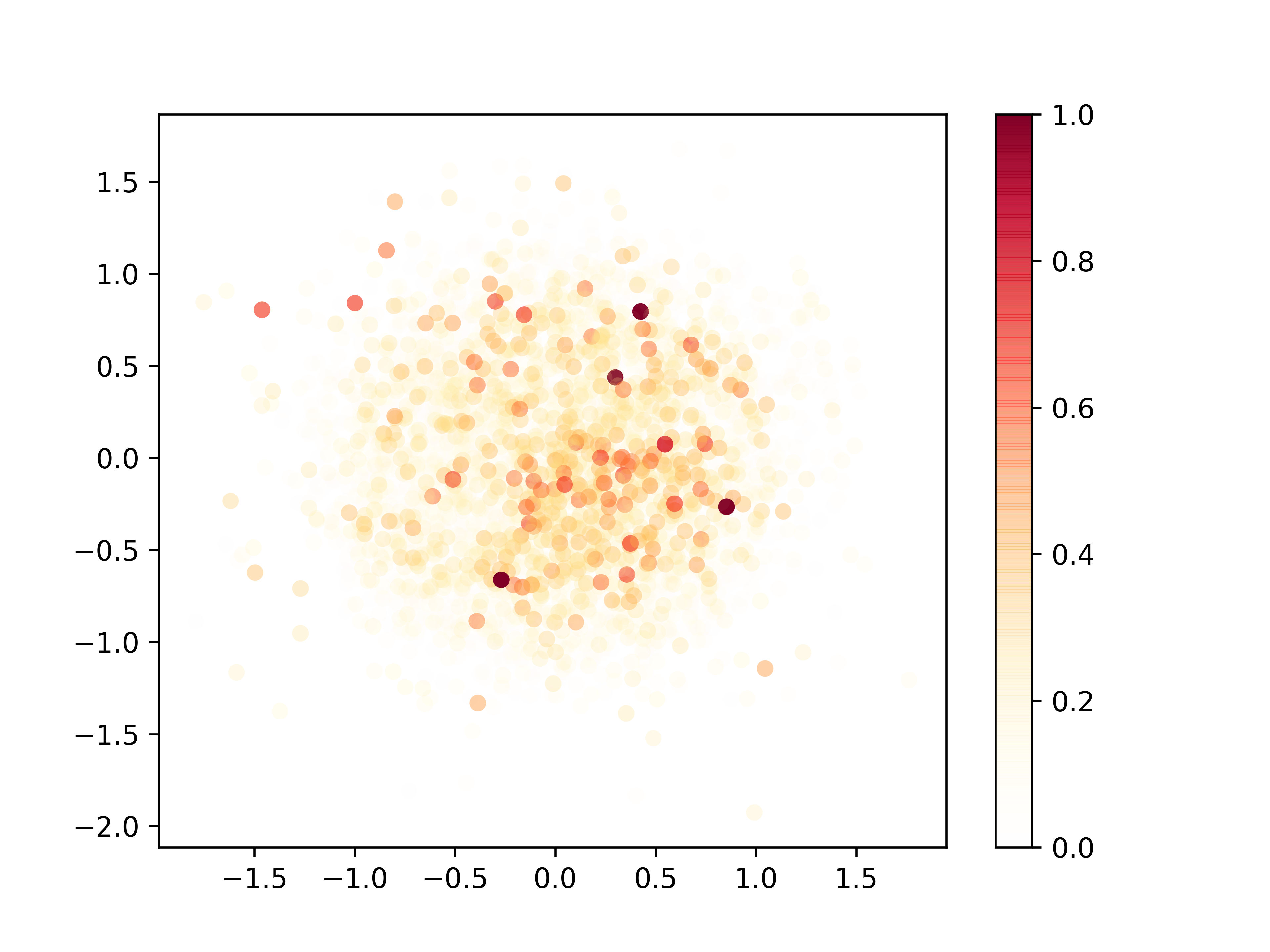}
\label{fig:huevos3}}
\caption{Multidimensional Scaling (MDS) plots at the beginning (a), middle (b) and end (c) of 
the training process (CrGAN-Cnet). Points' coordinates are obtained by applying MDS on 
real and synthetic data transformed by the critic's $h$ function. Left: real points (blue) and synthetic (red). Right: color 
intensity represents local discrepancy (contribution to the JSD divergence). }
\label{fig:huevos}
\vskip -0.1in
\end{figure}

In Figure \ref{fig:huevos}, we show the multivariate divergence estimates (JSD) 
and its point-wise decomposition at 
three stages of the training process (beginning, middle and end) for our method. The estimations were calculated using 12000 points from each set. To further reduce the variance, the results of ten runs were averaged. We observe how the matching of the real and generated distributions improves as the training progresses.

Moreover, we train classifiers to  separate real from generated data for the 
proposed CrGAN-Cnet method. The training set contains equal number of real and 
synthetic samples, and we evaluate on a test set that also contains an equal 
number of real/synthetic samples. A simple Logistic Regression with L2 
regularization classifier obtains an average accuracy of 58$\%$. However, a 
more powerful Random Forest (RF) classifier is able to separate the two sets 
better, with an average accuracy of 69$\%$. 
 Nevertheless, we are more interested in the ability of using the synthetic 
 data for training classification models for certain business problems, and as 
 we will show, the quality of the generated data is good enough for these 
 applications.

\begin{figure}[ht]
\vskip 0.1in
\centering
\includegraphics[scale=1.05]{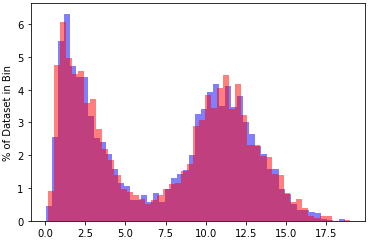}
\caption{Histogram of the distances between the generated samples (CrGAN-Cnet) and the nearest neighbours in the training and test sets (training set in red, test set in blue).}
\label{fig:hist_NN}
\vskip 0.1in
\end{figure}

As described in Section \ref{sec:validation}, we test whether the GAN is memorizing the training data.
The histograms of the distances between the generated samples (CrGAN-Cnet) and their 
corresponding nearest neighbours in the training and test sets are presented in 
Figure \ref{fig:hist_NN}. We speculate that the observed bi-modality is due to the discriminating purpose of the non-linear critic function $h$ used to transform the data. One can see that both distributions are very 
similar. Moreover, the KS test yields a $p$-value of 0.96 and the Bayesian 
Wilcoxon test accepts the null hypothesis (with posterior probability in the
[0.235,0.245] interval). This allows us to assert that the generative model is not simply memorizing the 
training set but rather learning its distribution.

We are particularity interested in using the GAN model to generate realistic data that we can legally use for our applications. In particular, we want to train models on synthetic data, to later apply on real data. To determine whether this is feasible, we train RF models  on training sets containing either only real or only synthetic samples generated by our model. We then test the model's performance of an evaluation set consisting of only real samples. By comparing the performances of the two models, we can determine if the generated samples are good enough to act as a proxy for real data. Results are presented in Table \ref{tbl:crossEval_bl}. One can appreciate that although the performance degrades slightly, the models trained on synthetic data are able to successfully classify the real samples for both classification problems.

\begin{table}
 \caption{RF classifier performance for Business/Leisure (BL) and Nationality (Nat) predictions, trained on real or synthetic set (CrGAN-Cnet), both evaluated on a real test set.}
  \label{tbl:crossEval_bl}
   \begin{center}
  \begin{small}
\begin{sc}
  \begin{tabular}{lcr}
    \toprule
  Method                   & BL Avg. Acc & Nat Avg. Acc \\
    \midrule
Real      &   0.94  & 0.78  \\
Synth &   0.92   &  0.71 \\
  \bottomrule
\end{tabular}
\end{sc}
\end{small}
\end{center}
\vskip -0.1in
\end{table}


Finally, we compare the performance of our approach against the methods described in Section \ref{sec:validation}. Similarly to the results presented before, Table \ref{tbl:rf_perf_others} shows the performance of RF classifiers trained to separate real and generated data produced by the different methods. One can appreciate that our method outperforms the other ones. Results show that using Cross-Nets in conjunction with fully connected layers improves the quality of the generated data, that the  Cram\'{e}r GANs methods outperform the Wasserstein GANs ones, and that using a simple numerical encoding of the categorical features performs significantly poorer than the embedding based approach. It should be noted that we do not claim that  Cram\'{e}r GANs always outperform WGANs. Nevertheless, based on our observations, Cram\'{e}r GANs generate better quality data than WGANs for our application.

\begin{table}
\caption{RF classifier performance separating real and generated data produced by the different methods. The optimal value would be 0.5.}
  \label{tbl:rf_perf_others}
   \begin{center}
  \begin{small}
\begin{sc}
  \begin{tabular}{lr}
    \toprule
  Method   & Avg. Acc.\\
    \midrule
CrGAN-Cnet & 0.69\\  
CrGAN-FC & 0.73\\   
WGAN-FC  & 0.75\\  
CrGAN-Num &  0.89\\
WGAN-Num & 0.93 \\
  \bottomrule
\end{tabular}
\end{sc}
\end{small}
\end{center}
\vskip -0.1in
\end{table}


\section{Conclusions and Future Work}
\label{sec:conclusions}

We propose a method to generate synthetic Personal Name Records (PNR) based on Generative Adversarial Networks (GAN).
The model takes a set of real PNRs made up of both numerical and categorical features as input, and uses them to train a Cram\'{e}r GAN.
The trained model is able to produce synthetic PNRs that follow the original 
data distribution without memorizing.

The method was tested on a real PNR dataset. The generated data was validated 
using several evaluation methods, including a recent point-wise Jensen-Shannon 
divergence decomposition that could be used in other application domains to 
evaluate or filter GAN generated data. In addition, we have tested the 
synthetic data by using it to train classification models associated to two 
important business applications in the travel industry. 

Experimental results 
show that the generative model is able to produce realistic synthetic data that 
matches well the real PNR distribution and that can be used for the selected 
business cases. Moreover, the proposed method outperforms other tested 
approaches.

In the future, we would like to explore the use of Conditional GANs to produce data conditioned on user inputs (e.g. generate French passengers departing from Paris in a month). Furthermore,  we would like to determine if a similar approach could be applied to other travel related data such as airline itinerary choice data \cite{choiceModel}.



\bibliography{yo2}
\bibliographystyle{icml2018}

\end{document}